\documentclass[journal]{IEEEtran}

\usepackage{times}
\usepackage{epsfig}
\usepackage{graphicx}
\usepackage{amsmath}
\usepackage{amssymb}
\usepackage{braket}
\usepackage{gensymb}
\usepackage{booktabs}

\usepackage{amssymb}% http://ctan.org/pkg/amssymb
\usepackage{pifont}% http://ctan.org/pkg/pifont

\newcommand{\cmark}{\ding{51}}%
\newcommand{\xmark}{\ding{55}}%
\usepackage{caption}
\usepackage{tabularx}
\usepackage{graphicx}
\usepackage{subcaption}
\usepackage{duckuments}

\newcommand{\hide}[1]{}

\usepackage[pagebackref=true,breaklinks=true,letterpaper=true,colorlinks,bookmarks=false]{hyperref}

\usepackage{orcidlink}

\begin{document}
\title{HoloMine: A Synthetic Dataset for Buried Landmines Recognition using Microwave Holographic Imaging}
\author{Emanuele~Vivoli\orcidlink{0000-0002-9971-8738},~\IEEEmembership{Member,~IEEE,}
        Lorenzo~Capineri\orcidlink{},~\IEEEmembership{Senior Member,~IEEE,}
        and~Marco~Bertini\orcidlink{},~\IEEEmembership{Member,~IEEE}% <-this % stops a space

\thanks{E. Vivoli and M. Bertini are with the Media Integration and Communication Center, Viale Giovanni Battista Morgagni, 65, 50134 Firenze FI, Italy}% <-this % stops a space
\thanks{L. Capineri is with Ultrasound and Non-Destructive Testing Laboratory, University of Florence, Italy.}% <-this % stops a space
\thanks{e-mail: emanuele.vivoli@unifi.it.}
\thanks{Manuscript received July 9, 2024; revised ??, 2024.}}

\markboth{IEEE Journal of Selected Topics in Applied Earth Observations and Remote Sensing,~Vol.~xx, No.~x, Month~YEAR}%
{Shell \MakeLowercase{\textit{et al.}}: Bare Demo of IEEEtran.cls for IEEE Journals}

\maketitle

\begin{abstract}
The detection and removal of landmines is a complex and risky task that requires advanced remote sensing techniques to reduce the risk for the professionals involved in this task.
In this paper, we propose a novel synthetic dataset for buried landmine detection to provide researchers with a valuable resource to observe, measure, locate, and address issues in landmine detection. The dataset consists of 41,800 microwave holographic images (2D) and their holographic inverted scans (3D) of different types of buried objects, including landmines, clutter, and pottery objects, and is collected by means of a microwave holography sensor.

We evaluate the performance of several state-of-the-art deep learning models trained on our synthetic dataset for various classification tasks. While the results do not yield yet high performances, showing the difficulty of the proposed task, we believe that our dataset has significant potential to drive progress in the field of landmine detection thanks to the accuracy and resolution obtainable using holographic radars.

To the best of our knowledge, our dataset is the first of its kind and will help drive further research on computer vision methods to automatize mine detection, with the overall goal of reducing the risks and the costs of the demining process.
\end{abstract}

\begin{IEEEkeywords}
Holographic imaging, Landmine detection, Robotic platform, Holographic dataset
\end{IEEEkeywords}

% \fi
\IEEEpeerreviewmaketitle

\section{Introduction}
Landmines are a significant threat to human life and the environment in many regions of the world \cite{landmine:report}, with several thousand persons killed or maimed every year. Their detection and removal is a complex and risky task that demands advanced remote sensing methods and technologies to reduce the risk of death or injuries for the professionals involved in this task; it is estimated that one deminer is killed and two injured for every 5,000 successfully removed mines \cite{khamis2016landmines}. Traditional detection methods involve detection with hand held instruments by human experts, the use of trained animals \cite{landmine:rats, review:sensing}, and the use of specialized equipment such as impulse ground-penetrating radar (GPR) \cite{gpr:dataset, gpr:cnn, gpr:faster-rcnn, gpr:autoencoder}, metal detectors \cite{landmine:metaldetector}, and thermal imaging cameras \cite{landmine:thermal}. Despite the considerable progress made in landmine detection methods, these approaches exhibit inherent limitations \cite{cambodian:website, Pochanin-2020}. For instance, plastic mines contain very small amounts of metal and are constructed using plastic materials to evade detection. As a result, the identification and clearance of buried landmines remain arduous and time-consuming tasks, incurring significant monetary costs. The United Nations has estimated that the removal of a single mine can cost between US \$300 and US \$1,000 \cite{Doswald-Beck-1995}.

%\hspace{-15mm}
\begin{figure}[!t]
    \centering
    \includegraphics[scale=0.20]{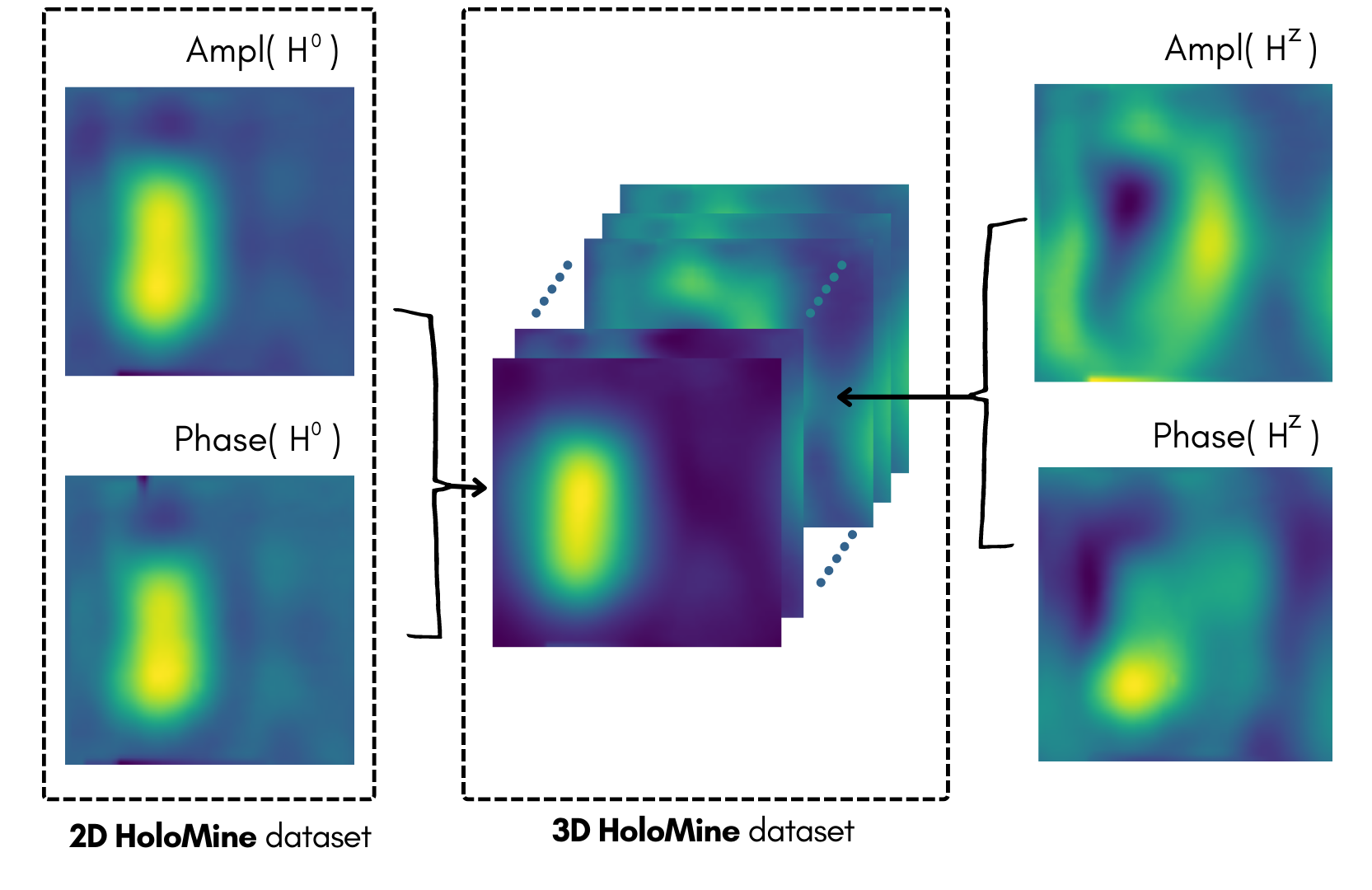}
    \caption{Detail of a PMN-4 scan represented in 2D complex holographic image $H^0$ and 3D reconstruction $\overline{H}$. On the right, is presented amplitude and phase of $H^z = f(H^0, z)$.}
    \label{fig:holo-2d-3d}
    \vspace{-10pt}
\end{figure}

On the other hand, microwave holographic radars have the ability to penetrate the soil up to 10-20 cm and be used to detect buried objects. Compared to GPR, holography is known for its precision and ability to generate high-resolution images of the shape and position of buried objects. 
Moreover, the visual outputs produced by holographic radars are simpler for operators to interpret compared to GPR images, which require specialized analysis skills. This advantage has led to the proposal of using holographic radars for mine detection. These holographic images are generated through the use of inversion algorithms \cite{goodman_introduction_2005} including methods such as the angular spectrum and Fresnel inversion \cite{schnars_digital_2005}. Such methods allow for the reconstruction of the shape and distance of buried objects by analyzing the electromagnetic reflectivity of objects in a plane parallel to the scanning plane. This enables the reflective scene picked up by the radar to be reconstructed in reverse along with the radar image.

However, creating a dataset for buried landmine detection is a time-consuming task, requiring scanning the ground some days after burying the landmines. Moreover, knowing the exact position of the buried object can be challenging \cite{zhuravlev_holographic_2013} since the soil is subject to change with time and weather conditions, potentially altering the objects' position w.r.t.~where they were originally buried. As a result, there is currently no common dataset available for either GPR data or holographic data.

To overcome the challenges posed by creating a dataset, we propose a synthetic dataset consisting of microwave holographic images of buried objects for classification and localization. 
Creating a synthetic hologram is a faster and more  efficient process than physically burying an object and scanning it in the ground. Moreover, synthetic holograms can be easily manipulated and customized to simulate different scenarios and conditions, which is not possible with physical objects in the ground. Despite the possible differences in environmental conditions (i.e. temperature and humidity) and variations in object-ground interactions, our proposed dataset ``HoloMine" aims to address these concerns by utilizing a rigorous process to create realistic holograms, obtained combining holograms of real objects and real ground scenes, that accurately simulate buried objects.
The dataset includes 2D and 3D images (as shown in Fig. \ref{fig:holo-2d-3d}), of different types of objects in multiple scenarios; we used a set of 6 military-grade replicas of mines used to train deminers, that replicate all the physical features of real mines, along with several types of clutter and objects. Our approach consists of collecting 200 real-ground scans and 208 in-air scans of objects to create the 41,800 size dataset which, to the best of our knowledge, is the largest (and first) synthetic dataset publicly available.

To evaluate our synthetic dataset, we tested state-of-the-art baseline models for 2D and 3D data. Our results indicate that there is significant room for improvement for these models in performing the classification tasks. However, the problem is difficult even for trained experts. Therefore, while the limitations of synthetic holography should be considered, we believe that our dataset is a valuable contribution to the field of buried landmine detection research for the computer vision community and can provide an accurate and realistic representation of buried objects in various scenarios.

To summarize, the key contributions of our work are:
\begin{itemize}
    \item to recognize the difficulties in creating a real dataset due to the uncertainty of the object's position and the time required for the object-ground stabilization;
    \item to pinpoint a new approach for this task thanks to the properties of holograms and holography inversions which allowed us to fuse scans and obtain the biggest synthetic dataset for buried landmines and clutters;
    \item to carry out extensive experimentation and providing baseline results on the provided dataset.
\end{itemize}

Data, baselines and pre-trained models are available at the link \url{https://github.com/emanuelevivoli/asmara}\footnote{available upon acceptance.}.
\section{Related works}
% RELWORKS-0 landmine clearance
\subsection*{GPR and landmine detection}
% gpr
Ground-penetrating radar (GPR) \cite{gpr} is one of the most commonly used techniques for detecting large buried objects such as pipes and walls \cite{ekes2012robot}, and small objects such as stones, other material objects, and landmine \cite{gpr:cnn, gpr:autoencoder, gpr:faster-rcnn}. % gpr for landmine
Among the works that tackle the task of landmine recognition \cite{57c288de48584debbc61040d0ef0697f}, the vast majority uses a simulator \cite{gprMax} to generate GPR Bscans with different soil composition and buried objects shape and material. In \cite{1427776} the authors propose a neural network that combines a complex-valued layer, to deal with the scattering parameters measured by GPR, with Self Organizing Map. Only few works use deep learning for classification or localization \cite{gpr:cnn, gpr:autoencoder, gpr:faster-rcnn, gpr:faster-rcnn-2, 1427776}. In particular, the authors of \cite{gpr:numerical} presented a two-step classification process which first isolate landmines from false alarms, and then recognize the type of each landmine; the authors trained their model using simulated 2D Bscans images. In \cite{gpr:cnn} the authors employed a CNN with pre-defined convolutional filters which allowed them to obtain high accuracy over a perfectly balanced generated dataset of 100k 2D patches in total. Despite the majority of the works being on the detection step of landmine clearance, two recent works \cite{gpr:faster-rcnn, gpr:faster-rcnn-2} approached the task of localization employing modern neural network architectures, such as Faster R-CNN on simulated and real data. 
However, these works have addressed the problem of distinguishing anti-tank (AT) landmine versus all other possible objects such as stones, woods and anti-personnel (AP) landmine. Moreover, when building the experimental setting, the authors considered only metallic AT and AP. In fact, GPR has some limitations such as the inability to detect very shallow buried objects, which may require post-processing to generate high-resolution plane images. Additionally, the technology may have difficulty detecting small non-metallic objects.
\subsection*{Holography and Inversion algorithms}
% holography
As an alternative to GPR, \cite{bossi-2022} developed an HSR system (Holographic Subsurface Radar), which uses a circular waveguide antenna with a single feed. The antenna operates at $1.9$ GHz. The choice of this frequency for the holographic radar is based on an in-depth study of the electromagnetic characteristics of the soil of Ukraine, which today represents an important application scenario for humanitarian demining~\cite{Bechtel2018TerrainAI}. Additionally, holographic radars are low-cost with the possibility of using 3D printing for their production \cite{bossi-2022}. With this setting, the holographic radar has a theoretical resolution equal to a quarter of the wavelength, i.e., about $15 cm$ in the air and about $3 cm$ in the target ground \cite{rascan}, and makes it possible to use holographic radars in the detection of landmines \cite{landmine:proceeding}. This value is a compromise between penetration depth and resolution as reported in the paper. In this way, the images of the data set contain information on the size and shape (circular or elongated artifacts can be thus discriminated). A series of published works show how the designed holographic radar provides such information \cite{Borgioli2018AHR,Razevig2019InfluenceOE}. Finally, the design of the holographic radar has been addressed to obtain the maximum sensitivity to the dielectric contrast that allows to reconstruct images of weakly reflecting anti-personnel mines (Low Metal Content AP landmines), and the design of the antenna and electronics is reported in \cite{Bossi2022VersatileEF}.
% Differently from the GPR, Holography uses continuous waves transmission.
One convenient property of holography is the holographic image reconstruction, arguably the most effective processing algorithm for microwave landmine detection. With the same principle of optical holography \cite{microwave:light}, through image reconstruction algorithms it is possible to obtain 3D images from 2D holograms, revealing objects' size and location.
Despite being the Fresnel transform algorithm the most commonly used method for holographic inversion, its limitations, such as sensitivity to phase errors and inability to deal with high frequencies, make it unusable in this context. However, \textit{Sheen et al.} \cite{angular-spectrum:sheen}, inspired by a reconstruction algorithm originally derived from microwave holography techniques \cite{microwave:holo}, designed a modification able to deal with high-frequency data and being less sensitive to phase errors. Our approach considers both 2D holograms and angular spectrum inverted holograms (3D) to provide a comprehensive dataset for landmine detection.

\subsection*{Landmine datasets}
Buried landmine datasets are needed to develop methods for automatic detection and recognition of mines. For this purpose, mine replicas that mimic real-world mines and are designed for training deminers are commercially available, providing a safe way to generate such datasets. However, the creation of such datasets is extremely impractical.

The approach of locating an object under the ground in a specific position and considering it to stay fixed also after weather agents and time is not realistic due to the fact that objects usually move in the soil, caused by varying conditions such as humidity, vibrations, etc. To approach this issue, \textit{Counts et al.} \cite{gpr:dataset} simulated the terrain using sandpits, inserting the objects knowing their spatial position in all 3 dimensions. This is possible because, differently from real soil, there is no need to wait for atmospheric agents for the sand to settle. Authors created the first publicly released dataset composed by a sandbox and many superficial and subsurface targets \cite{gpr:dataset:website}. To make the scans, a GPR was attached to a 3D mechanical positioning system located at a distance over a sandbox. The set of air targets and subsurface targets comprises metallic AP and AT mines, rocks, metal spheres, and corrugated pipes. Despite the effort in creating the dataset, using sand and metallic objects limits the applicability to different soil composition terrains and to plastic or low-metallic landmines. In order to preserve precise spatial localization and address different soil composition, later works are mostly based on gprMax simulations \cite{gprMax} that allow them to create custom dataset considering various objects and soil dielectric properties. Using simulations can have positive aspects. In fact, it makes the setting completely reproducible, and also the 3D location of the objects are known which can be suitable not only for detection but also for localization tasks \cite{gpr:faster-rcnn, gpr:faster-rcnn-2}. However, the simulations divert the effort from generalizing on different terrains and objects to terrains that can be reproduced by the simulator and objects that the simulator considers (e.g., regular shapes such as cylinders, cubes, and cones). Furthermore, the simulator may suffer from programming errors and approximations that would bias the algorithms to simulator artefacts that are not present in nature and difficult to reveal.

On the other hand, when considering holograms, their main application is in optical holography, where numerous works and datasets are available \cite{microwave:light,gabor:1948}. In the domain of landmine detection, the application of holography has been limited to a few studies, and at present, there exists no dataset specifically tailored to buried object detection utilizing holographic techniques. From the experiences of previous works, our approach is to create a new dataset for holographic subsurface radars that maintains the veracity of the data and avoids using unrealistic simulators and experimental settings. In so doing, similarly to \cite{gpr:dataset}, we developed a 3D-positioning structure together with an HSR and used them for acquiring different scans of outdoor and indoor elements; then, we merged these holographic images.

\begin{figure}[!htb]
    \centering
    \includegraphics[width=\columnwidth]{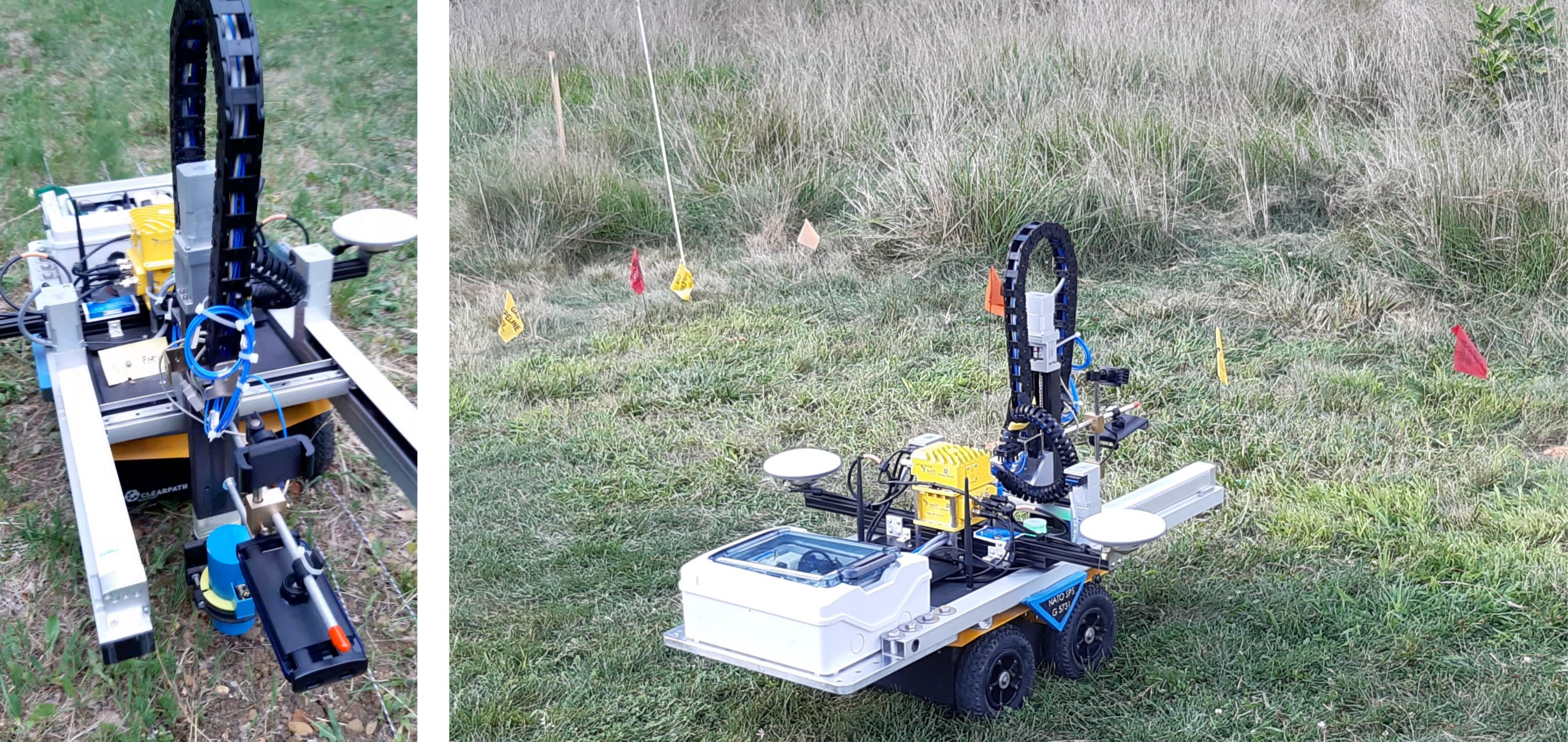}
    \caption{\textit{left)} view of the 3-D mechanical arm and Holographic Subsurface Radar (HSR); \textit{right)} a panoramic view of the Robotic platform. This system has been used to acquire the basic elements used to create the dataset.}
    \label{fig:ugo-first}
    %\vspace{-10pt}
\end{figure}

\section{Methods}
In this section, we present our novel holographic dataset for buried objects. We start by describing the acquisition setup (Fig.~\ref{fig:ugo-first}) and the collected signals (Fig. \ref{fig:transform}), then we dig into the pre-processing and the fusion procedure to obtain the 2D holograms and 3D inverted holograms, and the dataset generation pipeline (Fig. \ref{fig:experimental-setup}).

% METH-0
\subsection*{Radar setup}
\begin{figure}[t]
    \centering
    \begin{subfigure}{0.45\textwidth}
        \hspace{-10mm}
        \includegraphics[scale=0.49]{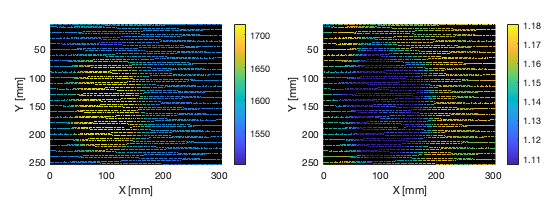}
        \caption{Scattered time-series of $amp(S)$ [left] and $phase(S)$ [right] based on positions $P$.}
        \label{fig:transform:zig-zag}
      \end{subfigure}
      \begin{subfigure}{0.45\textwidth}
        \hspace{-10mm}
        \includegraphics[scale=0.49]{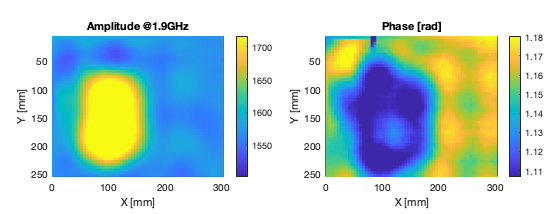}
        \caption{Amplitude $amp(H)$ [left] and Phase $phase(H)$.}
        \label{fig:transform:result}
      \end{subfigure}
    \caption{Transformation from zig-zag time-serie acquisition $S,P$ to $H$ through $T:S,P \rightarrow H$.}
    \label{fig:transform}
    \vspace{-10pt}
\end{figure}

The acquisition setup, based on \cite{bossi_design_2020} - the system is shown in Fig.~\ref{fig:ugo-first}, consists of a 3-D mechanical movement and a Holographic Subsurface Radar that are controlled through the Robot Operating System (ROS). The radar operates at $1.9$ GHz and it is mounted on the positioner support, which, in less than $3$ minutes, covers a surface of $30\times30$ cm, capturing the signal every $100$ ms. We collect in a zig-zag scan a list of two-value signals, amplitude, and phase respectively, meaning the ensemble of reflections of soil and buried objects in a particular point of the positioner.  Fig.~\ref{fig:transform:zig-zag} shows scattered amplitude and phase collected in the first stage. In order to represent the time-series signals $S(t)$ and the positioning information $P(t)$ in a single spatial dimension image $H$, a transformation $T:S,P \rightarrow H$ is needed. In particular, the transformation corresponds to a spatial interpolation (among $P$) of real and imaginary components of the signal $S$. The resulting image is a complex image of $60\times60$ pixels so that $H \in \mathbb{C}^{60\times60}$, and every pixel is located at $0,5cm$ distance; amplitude and phase elements of $H$ are shown in Fig.~\ref{fig:transform:result}. 

\subsection*{Acquisition criteria}
\begin{figure*}[t]
    \centering
    \includegraphics[scale=0.3]{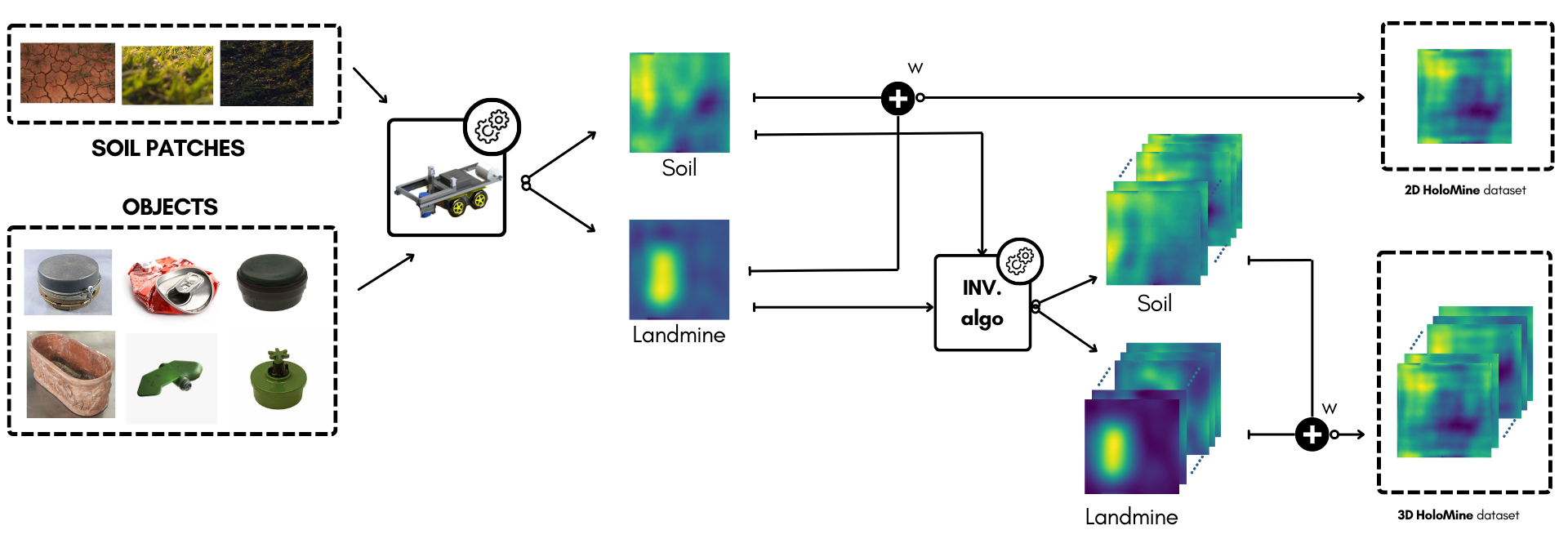}
    \caption{Dataset creation pipeline from indoor and outdoor scan acquisition [left] to pre-processing [middle] to 2D and 3D image creation [right]. $w$ indicates the weight used to fuse the hologram images.}
    \label{fig:experimental-setup}
    \vspace{-10pt}
\end{figure*}

Inspired by \cite{bg-sub-1, bg-sub-2, bg-sub-3, gpr:surface-effect}, our approach consists of acquiring two sets of data: outdoor $H^{OUT}$ and indoor $H^{IN}$, and then mixing these scans with the pipeline shown in Fig. \ref{fig:experimental-setup}. Firstly, a collection of $50$ terrain patches was considered, in specific atmospheric conditions, in which the absence of particular objects is ensured (i.e.~mines and clutter). Every square patch $H^{out}_{j, d_o}$ is scanned from the four cardinal directions (\textit{North, South, West, East}), forming a set of 200 scans as follow:

\begin{equation}
  H^{OUT} = \Set{ H^{out}_{j,d_o}  | \begin{array}{l}
    d_o \in \{N,S,W,E\}; \\
    j=1,\dots,50
  \end{array}}
\end{equation}

Secondly, airborne scans in a laboratory environment of 13 objects $o_k$ (including AP landmines, pottery objects and clutters). Every object was located both at distance $h = 4 \text{cm}$ and $h = 8 \text{cm}$ from the source. The object is oriented to each of the cardinal orientations $d_i \in \{N,S,W,E\}$ and is placed with a slope $s$ of $0\degree$ or $20 \degree$ in the $d_i$ direction. To be noticed that for indoor objects the orientation is $d_i$ while $d_o$ for outdoor objects. Let $H^{IN}$ being the set of all possible indoor scans, it is composed by all possible 16 configurations of every one of the 13 $o_k$ objects, creating 208 total indoor scans as follow:

\begin{equation}
  H^{IN} = \Set{ H^{in}_{k, h, d_i, s} | \begin{array}{l}
    k =1,\dots,13; \\
    h \in\{4;8\}; \\
    d_i \in\{N,S,W,E\}; \\
    s \in\{0;20\}
  \end{array}} 
\end{equation}

The implemented system provided us with precise control over object locations, including their coordinates and slope, while also providing knowledge of the material properties of the object.

During the scanning process, several measures were implemented to ensure accuracy. For indoor scans, the mobile platform was raised at least 1 meter above the ground to minimize interference from ground reflections. Additionally, care was taken to maintain a safe distance from reflective objects such as metallic furniture and cables in the laboratory. The 3D structure, mounted on an autonomous guidance system for automatic demining, was equipped with appropriate counterweights and oscillation damping systems to ensure minimal movement and clean signal reception by the HSR. In outdoor scans, proper calibration was required to ensure that the HSR did not come into contact with the ground, which could damage the radar and result in anomalies in the scans. Furthermore, the robot platform was secured firmly on all four wheels to prevent excessive oscillation on jagged terrain.

\subsection*{Hologram fusion}
In \cite{gpr:surface-effect}, one of the earliest works on buried object analysis using GPR, the authors study the effect of the height of the radar with respect to the surface. The surface depth profile is measured with a LIDAR and the effect on the acquisitions is modeled with a simulator in order to subtract it from the images, leading to positive effects on target detection. Following this approach, we aim at doing the opposite operation with consistent holograms: adding the HSR response of a $H^{IN}$ object acquired in air to a not homogeneous $H^{OUT}$ soil medium holographic scan. 

In this way, the two sets of $H^{OUT}$ and $H^{IN}$ can be used together to represent scenes in which a particular object configuration $H^{in}_{i, h, d_i, s} \in H^{IN}$ is located into the soil $H^{out}_{j,d_o} \in H^{OUT}$. However, the different permittivity of the air and soil media of the indoor and outdoor scans require to implement a weighting criterion. We employed a weighted sum of indoor and outdoor holograms using the following formula:

\begin{equation}
    H = \alpha \times H^{in} + (1 - \alpha) \times H^{out}
\end{equation}
  
where $\alpha$ determines the weight assigned to each hologram and must be chosen to ensure a realistic and balanced representation of both indoor and outdoor scenes in the final mixed dataset.

The $\alpha$ coefficient can be chosen based on two principle factors: (i) the dielectric permittivity differences between the air and the soil composition that affect the attenuation of microwaves; (ii) empirical analysis. In our work we have determined the best coefficient using the latter approach, which will be shown in the next section.

\subsection*{Inversion theory}

An important consideration noted in \cite{gpr:surface-effect} is that using the backprojection algorithm \cite{gpr:backprojection} helps to mitigate surface effects. This algorithm shares the same principles as the angular spectrum method phase reconstruction for holographic data.
From this knowledge, and inspired from \cite{angular-spectrum:sheen}, we adopted the angular spectrum algorithm to mitigate surface noise. In the system configuration, the source is located at $z=0$ and the Hologram $H$ is considered as $H^0$. The target is characterized by a reflective function $H^z = f(H^0,z)$ which lead to the definition of the reflection perceived at the transceiver calculated by the superposition of each point on the target plane at distance plane $z$. 
The image reconstruction is numerically calculated by the following formula:
\begin{equation}
 f(H^0, z) = FT_{2D}^{-1}\left\{ FT_{2D}\left\{ H^0 \right\}e^{-jz\sqrt{k^2-k_x^2-k_y^2}} \right\}
\end{equation}
where $FT$ represents the Fourier Transform operation, $H^0$ is the hologram at $z=0$, $k_x$ and $k_y$ represent the spatial frequencies in the $x$ and $y$ directions, respectively, and $k_z$ represents the wavevector in the $z$ direction (i.e., the direction of propagation of the electromagnetic radiation). The output of the function is the complex field $f$ at a distance $z$ from the hologram $H^0$.

Following the above formulation, \cite{holo:book} defines a MATLAB algorithms of Angular Spectrum and Convolution reconstruction while \cite{delafuente,holopy} adopted the Python language in structuring an open source tool which performs digital holograms and light scattering. In our case, we use HoloPy \cite{holopy}, a tool that is able to forward propagating light from a scattering as well as backward propagating light from a digital hologram. This allows us to obtain $\overline{H} = f(H^0)_{z=0, \dots, d}$, where $\overline{H} \in \mathbb{C}^{60\times60\times d}$ and $d$ is the number of slices that we want to recreate. Similar to the approach we employed for holographic images, we calculated the inverted complex fields $\overline{H}$ and merged them by utilizing the same coefficient as for Holograms ($w = 0.17$). 

\subsection*{Buried objects}

\begin{figure*}[t]
    \centering
    \subfloat[PMN-4\label{fig:object1}]{%
        \includegraphics[width=0.13\textwidth]{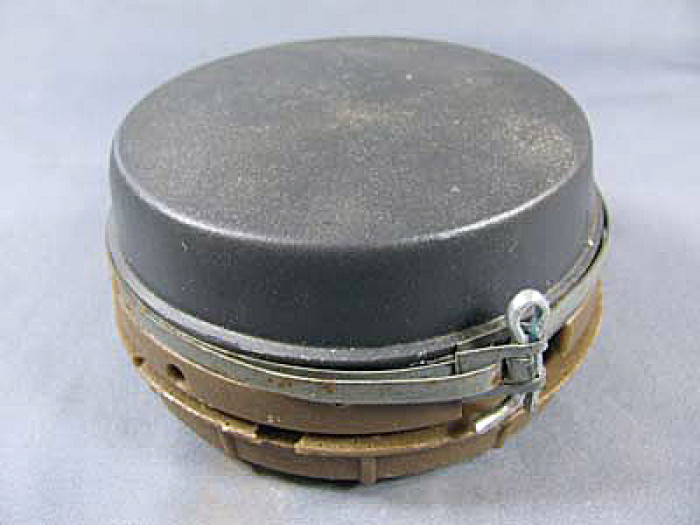}}
    \hfill
    \subfloat[PMN-1\label{fig:object2}]{%
        \includegraphics[width=0.13\textwidth]{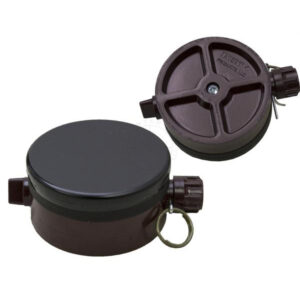}}
    \hfill
    \subfloat[VS-50\label{fig:object3}]{%
        \includegraphics[width=0.13\textwidth]{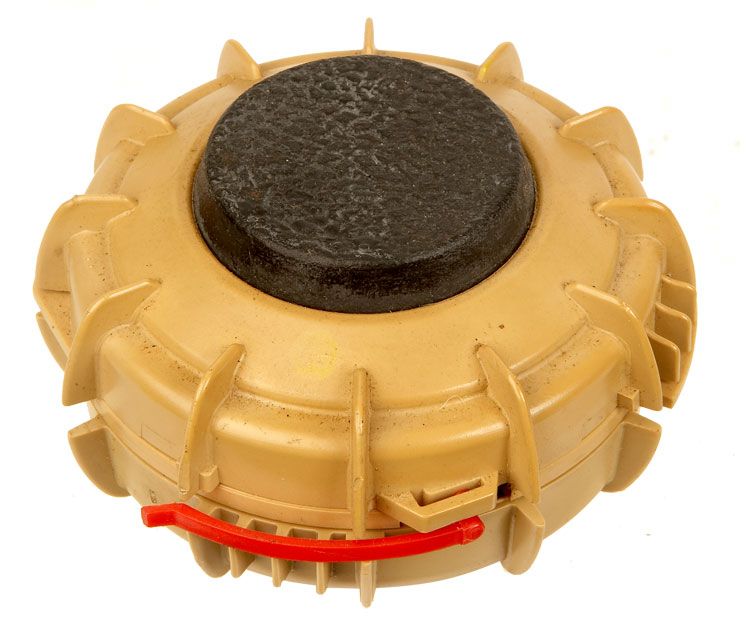}}
    \hfill
    \subfloat[TYPE 72\label{fig:object4}]{%
        \includegraphics[width=0.13\textwidth]{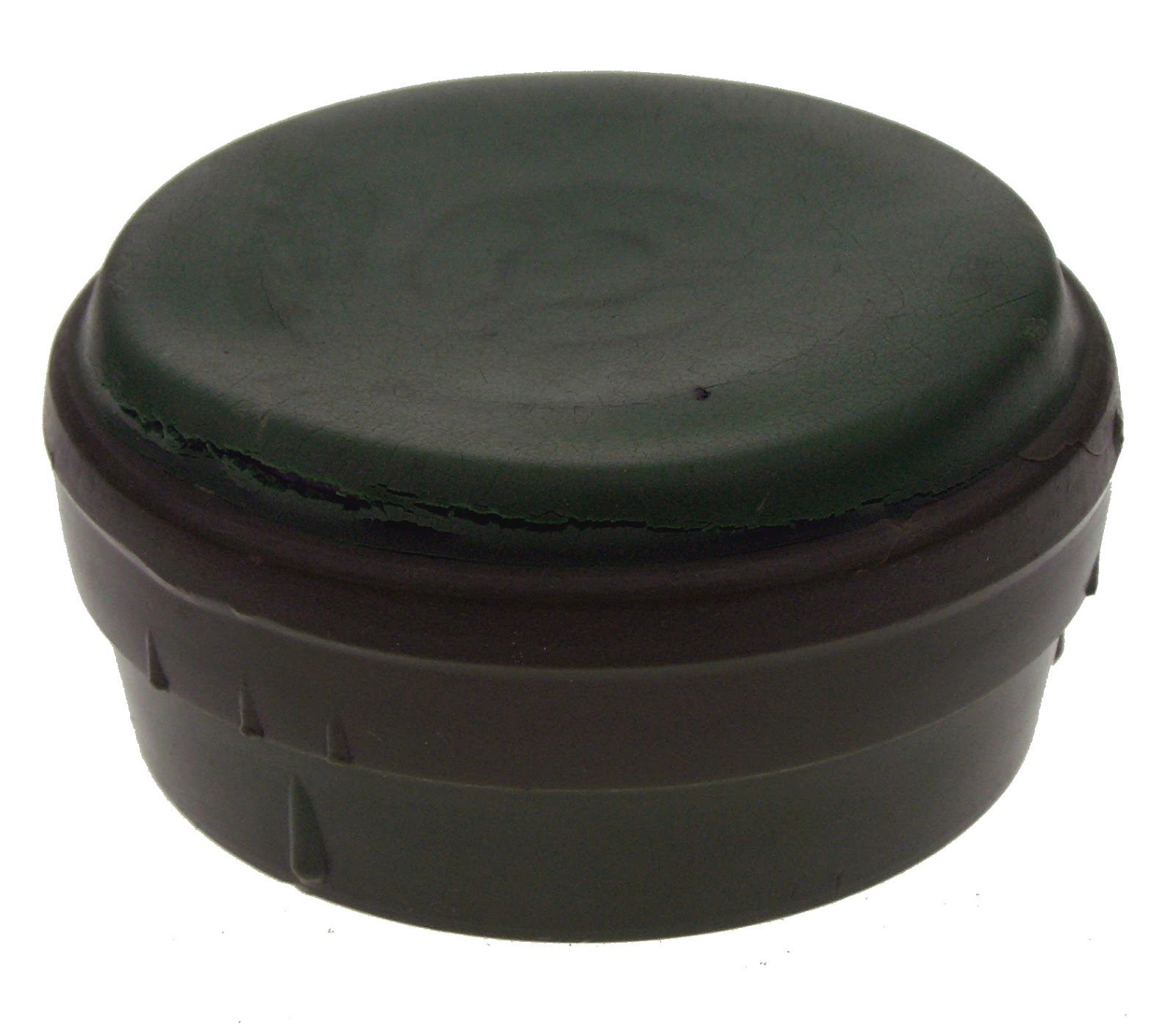}}
    \hfill
    \subfloat[M-14\label{fig:object5}]{%
        \includegraphics[width=0.13\textwidth]{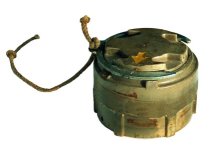}}
    \hfill
    \subfloat[PMA-2\label{fig:object6}]{%
        \includegraphics[width=0.13\textwidth]{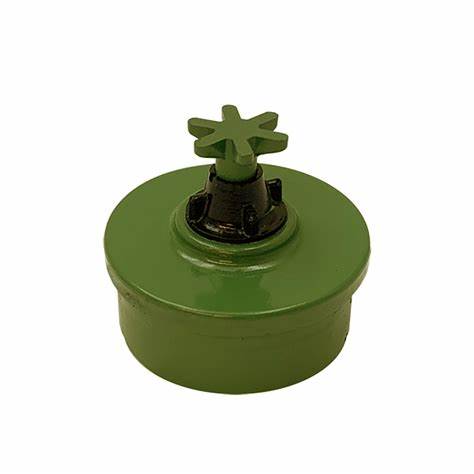}}
    \hfill
    \subfloat[Butterfly\label{fig:object7}]{%
        \includegraphics[width=0.13\textwidth]{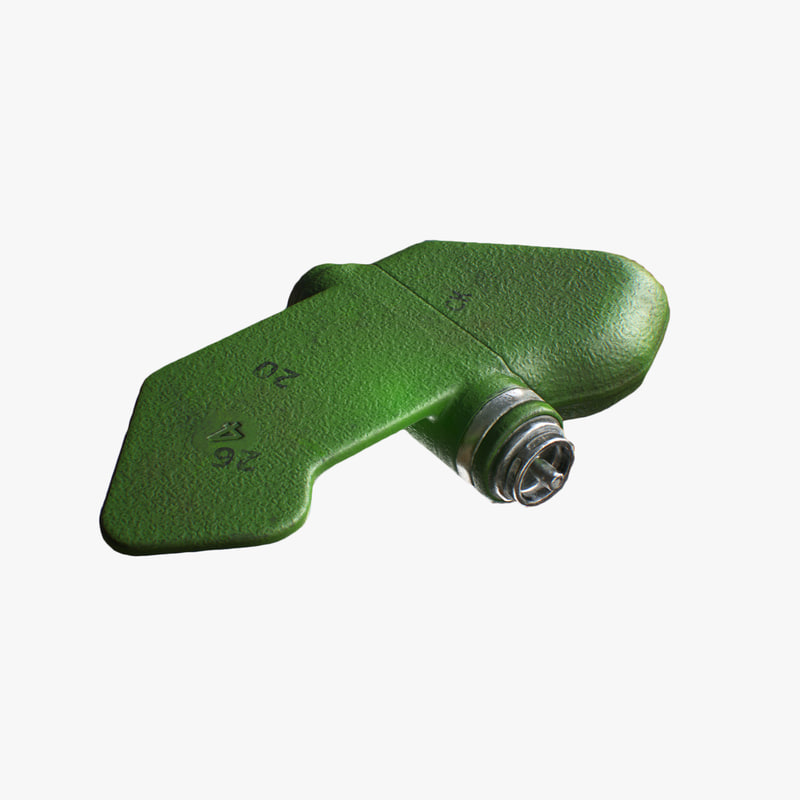}}
    \\
    \subfloat[Bullet\label{fig:object4}]{%
        \includegraphics[width=0.13\textwidth]{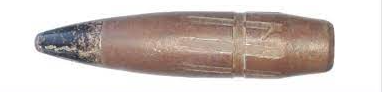}}
    \hfill
    \subfloat[Stone\label{fig:object8}]{%
        \includegraphics[width=0.13\textwidth]{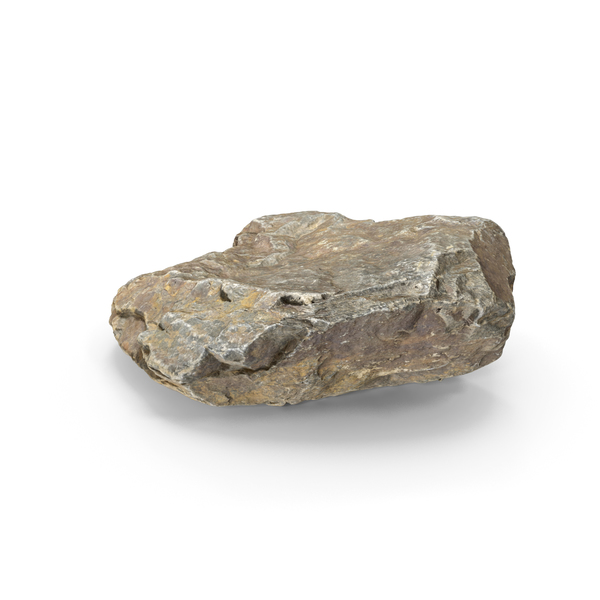}}
    \hfill
    \subfloat[Wood cylinder\label{fig:object9}]{%
        \includegraphics[width=0.13\textwidth]{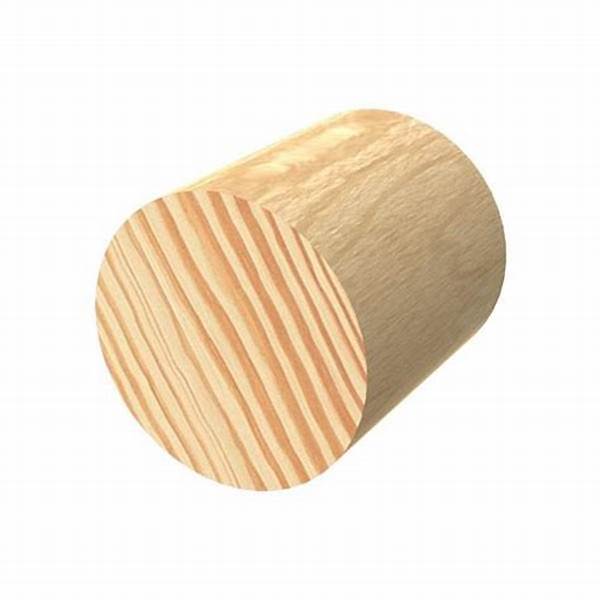}}
    \hfill
    \subfloat[Can\label{fig:object10}]{%
        \includegraphics[width=0.13\textwidth]{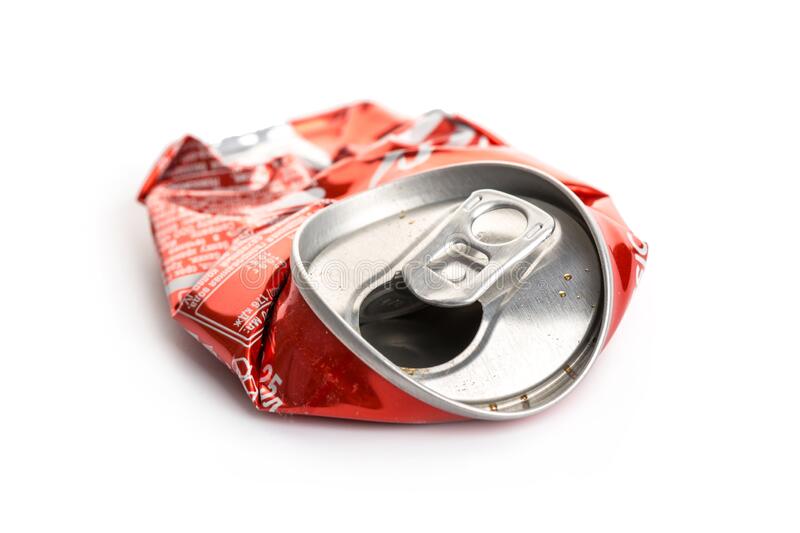}}
    \hfill
    \subfloat[Clay pot\label{fig:object11}]{%
        \includegraphics[width=0.13\textwidth]{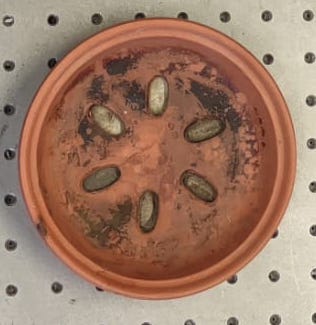}}
    \hfill
    \subfloat[Clay pot 2\label{fig:object12}]{%
        \includegraphics[width=0.13\textwidth]{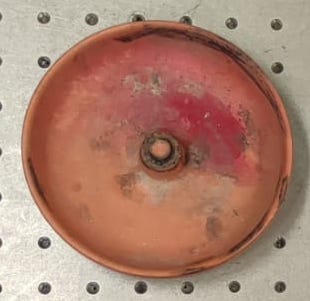}}
    \hfill
    \subfloat[Clay pot 3\label{fig:object13}]{%
        \includegraphics[width=0.13\textwidth]{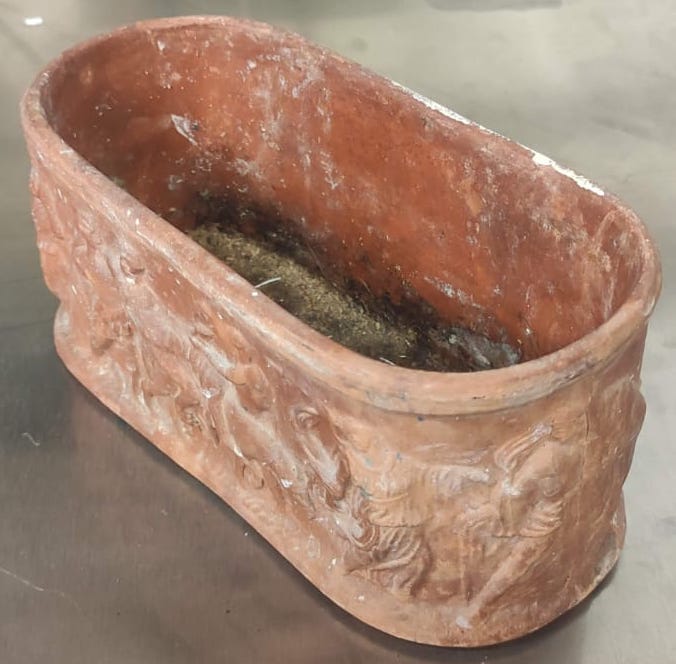}}

    \caption{Objects considered for indoor scans. On the first row, landmines replicas are reported, namely (from left to right): \textit{PMN-4} \textit{(d=95mm, h=46mm)}, \textit{PMN-1} \textit{(d=95mm, h=55mm)}, \textit{VS-50} \textit{(d=90mm, h=45mm)}, \textit{TYPE 72} \textit{(d=78mm, h=38mm)}, \textit{M-14} \textit{(d=56mm, h=40mm)}, \textit{PMA-2} \textit{(d=68mm, h=61mm)}, and \textit{Butterfly} \textit{(l1=112mm, l2=60mm, h=15mm).} The second row shows the \textit{Bullet} \textit{(d=20mm, l=120mm)}, some clutter (stone \textit{(l1=110mm, l2=56mm, h=34mm)}, wood cylinder \textit{(d=35mm, h=40mm)}, and crumpled can \textit{(l1=110mm, l2=62mm, h=15mm)}, and some pottery objects (perforated clay pot \textit{(d=180mm, h=28mm)}, clay pot \textit{(d=170mm, h=24mm)}, and deep clay pot \textit{(l1=120mm, l2=265mm, h=135mm)}).}
    \label{fig:indoor-objects}
\end{figure*}

The dataset has been carefully constructed with the inclusion of three distinct types of objects.
The first type consists of landmines objects (first row of Fig.~\ref{fig:indoor-objects}), which are highly present in Europe and East-Europe post-conflict countries, as well as in many other African and Asian countries. Their material composition vary for different electromechanical parameters as well as density, dielectric permittivity and metal conductivity. In particular, \textit{PMN-4} has a $1$cm metallic ring surrounding the hood while the \textit{Butterfly} landmine has no metallic concentration. Our choice to include also low-metal and plastic landmines (replicas) are due to the fact that plastics used to build them have low density and a non-uniform surface, making them more difficult to detect than metal objects; indeed, they have been designed to be hard or impossible to be detected by metal detectors.
The second type of clutter object included in the dataset are common clutters such as stones, wood and can, which are commonly observed in a post-conflict terrain (first three objects of second row in Fig. \ref{fig:indoor-objects}). Despite the particular shape of the considered objects, we are interested in the holographic response of materials different than plastic and metal.
Finally, the third type of clutter object included in the dataset is pottery, which are often present buried in the soil and can be mistaken for buried mines (last three objects of second row in Fig. \ref{fig:indoor-objects}). In fact, we included different types of pottery: shallow clay pot, a pot with holes and deep clay pot.
The inclusion of these three clutter object types in the dataset aims to create a diverse range of scenarios and conditions that can be encountered in real-world landmine detection tasks. This will allow researchers to develop and test algorithms that are robust and accurate in a variety of situations.

\subsection*{Data annotations and tasks}

\begin{table}
    \centering
    \resizebox{\linewidth}{!}{
    \begin{tabular}{l | c | cc | ccc}
        %         &           &            &            & \multicolumn{4}{c}{Task} &         \\ \cmidrule{5-8}
        Dataset                     &  \#scans  &  \#mines  & \#clutters    & CLS   & DET       & SEG   \\ \midrule
        \textit{Counts et al.} \cite{gpr:dataset}      & 14      &  7         & 4*             & \cmark & \cmark   & \xmark \\
        \textit{Giannakis et al.} \cite{gpr:numerical}      & 4k      & 2          & 5             & \cmark & \xmark   & \xmark \\
        % \textit{Malof et al.} \cite{DBLP:journals/tgrs/MalofRKFHWLCC19}                     & 13 $L$*      & 417k*         & 12            & \cmark & \cmark   & \cmark* \\
        \textit{Bestagini et al.} \cite{gpr:autoencoder}                      & 114        & 9           & -             & \cmark & \cmark   & \xmark \\ \midrule
         \textit{Kafedziski et al.} \cite{gpr:faster-rcnn}  &  48 $^{\dagger}$     & 9          &  -            & - & - & - \\
         \textit{Pham et al.} \cite{gpr:faster-rcnn-2} & 100 + 50 $^{\dagger}$   & -          & -             & - & - & -  \\ \midrule
        \textbf{HoloMine}                         & 41.8k       & 7           & 6            & \cmark & \cmark   & \cmark \\
        \bottomrule
        \multicolumn{7}{l}{{\small *in addition, dozens of stones were located on the surface.}}
    \end{tabular}}
    \caption{Comparison of HoloMine with related datasets. $^{\dagger}$ \ \ denotes the datasets are created with the gprMax simulator \cite{gprMax}. CLS (Classification), DET (Detection), and SEG (Segmentation) show which tasks the datasets support.}
    \label{tab:datasets}
\end{table}
Our dataset currently consists of both indoor and outdoor scans of various types of buried objects and soil, alongside their corresponding configuration details that facilitate easy data annotation for Recognition tasks. 
Recognition is a central task in computer vision that involves identifying and localizing objects of interest in an image or a video. It typically involves three sub-tasks: classification, detection, and segmentation. Classification refers to assigning a label or a category to an object or a region in an image. Detection involves localizing objects of interest and predicting their labels or categories. Segmentation is the task of partitioning an image into multiple regions or segments, where each segment corresponds to a distinct object or a part of an object.
In the context of landmine clearance, the tasks of detection, localization, and identification correspond to the sub-tasks of detection and classification in computer vision. Detection refers to the process of identifying the location of buried landmines in a given area, while localization involves accurately pinpointing the location of the landmines. Identification involves recognizing the type of the detected landmines.
Table \ref{tab:datasets} presents an inventory of datasets that incorporate GPR radars, which are either based on real-world measurements, simulations, or a combination of both. The table enumerates their structure and characteristics, including the number of scans, the number of different mine types, the types of clutter objects, and the tasks for which the annotations can be employed. Notably, our dataset is distinctive due to its significant size and its ability to facilitate comprehensive Recognition tasks for buried landmine detection, which includes classification, detection, and segmentation. It comprises both indoor and outdoor scans of different types of buried objects and soil, along with associated configuration details that facilitate simple data annotation. Furthermore, our dataset is exceptional in that it is not produced via simulations, unlike other similar datasets.

\begin{figure}[t]
    \centering
    \includegraphics[scale=0.22]{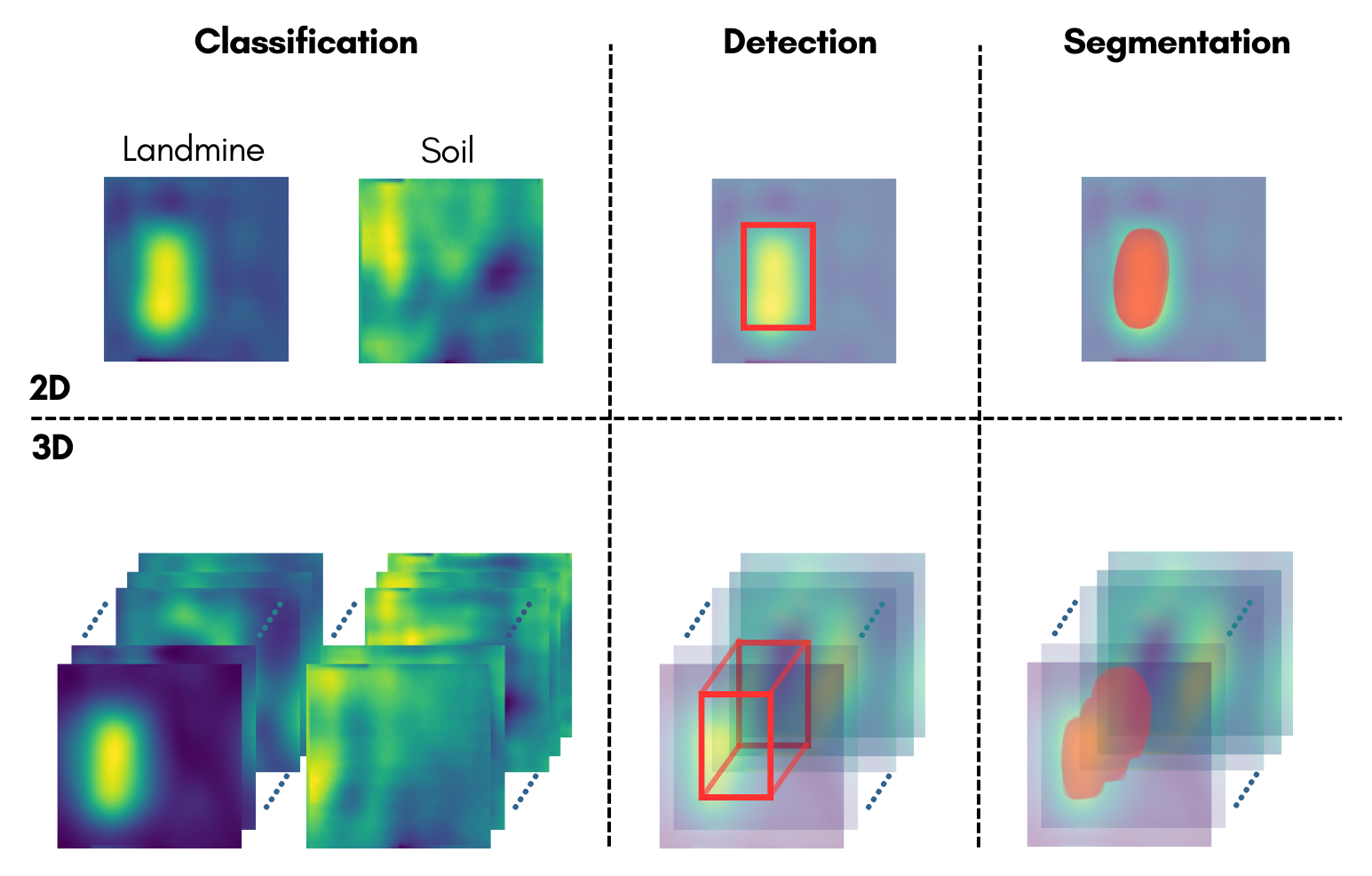}
    \caption{Representation of available labels for our HoloMine dataset. For visualization purposes, only the amplitudes $H^0$ and $H^z$ are shown.}
    \label{fig:annotations}
    \vspace{-10pt}
\end{figure}

To generate labels for the classification task, represented in Fig. \ref{fig:annotations}, we employed various techniques based on the complexity of the task. In fact, the classification task can be evaluated in terms of binary classification (mine/clutter), ternary classification (mine/clutter/pottery) or multi-class classification (a class for each type of buried object). For the binary classification task, where the classes are mine and non-mine, we selected 80\% mines and objects for the train-set and used the remaining objects for the test-set. By ensuring that no object could be present in both train and test sets, even from any of its 16 possible configurations, our dataset enables a more robust evaluation.
Nonetheless, in the case of soil patches, there was no distinction between terrain that was observed during training and during test process, since both landmines and clutter were embedded in each of the patches. This is possible because soil patches have a relatively homogeneous texture and appearance, making it difficult to distinguish one from another, which means that introducing different objects or clutter in different patches does not significantly affect the overall appearance of the soil patch.
To accomplish the three-class classification task, a similar labeling approach is employed. In this case, 80\% of the mines, clutters, and pottery objects are chosen for the train-set, and the rest of the objects in all of their configurations, in each of the 200 terrain scans, are reserved for the test-set. However, For the 13-class classification task, a different approach was used: we randomly split the data after performed the fusion of indoor and outdoor scans. To ensure a balanced distribution of objects across the train and test sets, we randomly split the data into 80\% train-set and 20\% test-set, while ensuring that all objects in a particular configuration were present in the same set.

To generate annotations for detection and segmentation, we used a combination of manual and automatic methods. For detection, we automatically labeled the position and size of the bounding boxes around each object in the 2D images since we have precise information about the location, dimensions, material, and shape of each object. We also provided a binary mask for each bounding box to indicate which pixels belong to the object and which do not. For segmentation, we used an automatic algorithm in CVAT \cite{CVAT_ai_Corporation_Computer_Vision_Annotation_2022} to generate initial segmentations of the objects. We then manually corrected any errors in the segmentations.

The annotations for both detection and segmentation are provided in a standard format for easy use with common deep learning frameworks.

\subsection*{Deep learning models}
In this work, we focus on deep learning methods for buried landmine detection, as they have shown promising results in various computer vision tasks. Specifically, we consider some of the most popular and widely used deep learning models, including ResNet \cite{he2016deep}, ViT \cite{DBLP:journals/corr/abs-2010-11929}, U-Net \cite{ronneberger2015u}, and EfficientNet \cite{tan2019efficientnet} for 2D images and ViT3D \cite{https://doi.org/10.48550/arxiv.2208.04309} and U-Net3D \cite{https://doi.org/10.48550/arxiv.1606.06650} for 3D images. These models have demonstrated excellent performance in various classification tasks and are thus well-suited to be used as baseline methods for the task of buried landmine detection.

To validate the effectiveness of our dataset and establish benchmark results, we trained and tested several deep learning models on the synthetic holographic data. Regarding 2D classifications, ResNet and EfficientNet are based on convolutional neural networks, while ViT is based on the transformer architecture that uses self-attention mechanisms, and U-Net is a popular encoder-decoder architecture used in medical image analysis tasks. As for 3D classifications, we have evaluated two models: ViT3D and U-Net3D. These models extend the ViT and U-Net models, respectively, to 3D data.

\textbf{Models modifications.}
Deep learning models are usually designed for specific input data sizes, such as images larger than 255 pixels per side and with 3 channels (RGB). In our case, we have a complex $60\times60$ image and we needed to decide how to use the available information. 
We chose to focus on selecting amplitude components and analyzing real 2D and 3D images, while leaving other approaches for future development discussed in the Future Work section.
Because we had a grayscale image, we had to modify the architecture of the models. In particular, ResNet, U-Net and EfficientNet have 3 input channels, so we adapted the first convolutional layer of the three models to accept grayscale input. Additionally, the last classification layer of ResNet had to be characterized to receive a smaller number of inputs due to the fact that the input images are significantly smaller. Another approach could be to maintain the same network size by adding padding to the input image. Despite adding zero-padding to the holograms is a common technique used to increase the field of view or the spatial resolution of the microwave hologram, adding too much zero-padding can lead to overfitting. In this case, the dimensions of cropped input images for ResNet50 $224\times224$, which means 13 times bigger than our images size, making padding unfeasible. Regarding the U-Net, since it is not designed for classification tasks but for segmentation tasks where the output has the same size as the input, we also had to add a classifier at the end of the network. 
In the experiments, we did not consider pre-trained networks. This is because we were interested in knowing the performance of the network not due to prior knowledge obtained from other data. Furthermore, although it is easy to find pre-trained ResNet, it is not as easy with 3D models. In the supplementary material extensive analysis on training procedure will be discussed.

\textbf{Metrics.}
For landmine detection, it is generally better to prioritize recall over precision. This is because the consequences of missing a landmine (false negative) are much more severe than falsely detecting a non-landmine object (false positive). In other words, it is more important to detect as many landmines as possible, even if that means some false positives than to miss any landmines. However, it is still important to strive for a balance between recall and precision to minimize both false positives and false negatives.  Based on this, for evaluating our benchmarks, we choose the $F_{1}$ metric for classification tasks. $F_{1}$ is a combination of precision and recall, which balances the trade-off between them:
\begin{equation}
F_{1} = 2 \cdot \frac{\text{precision} \cdot \text{recall}}{\text{precision} + \text{recall}}
\end{equation}
This choice of metric is appropriate for our dataset and task, where we need to balance the accuracy of identifying buried objects with the ability to correctly identify the absence of such objects. We use the Torchmetrics implementation using the specific binary and multiclass functions.

\section{Experiments}
In this section, we will present both the coefficient optimization experiments and the deep learning model experiments on 2D and 3D holographic images. 

\subsection{Coefficient Optimization for Hologram Fusion}
Our approach to landmine holographic imaging is to fuse synthetic in-air holograms $H^{in}$ and outdoor scans $H^{out}$, using a properly designed coefficient. One option is based on the dielectric permittivity of the air and soil. The dielectric permittivity of the air is $E_{r_{air}} \simeq 1$ and the estimation of the dielectric permittivity of the soil in the real field, with medium-high moisture content $E_{r_{soil}}\simeq 6$. This way, apart from a scaling factor $c$ that is applied to every resulting hologram, the function we consider is:
\begin{equation}
    H  = \alpha \times H^{in} + (1 - \alpha) \times H^{out} \\
    \approx \frac{\alpha}{(1-\alpha)} H^{in} + H^{out}
\end{equation}
where $\frac{\alpha}{1-\alpha}$ can be posed as the same ratio between $\frac{E_{r_{air}}}{E_{r_{soil}}} = \frac{1}{6}$. Solving this simple linear equation we obtain the $\alpha$ to be $\alpha = \frac{1}{7} = 0.143$.

In addition to this, we also perform an empirical study to estimate $\alpha$. We have collected a number of experimental holographic scans of the same objects from different perspectives. Intuitively, the holograms obtained from the same object in the same soil patch, but from a slightly different perspective, should be more similar than the scans of other objects. 

\begin{figure}[!htb]
    \hspace{-5mm}
    \includegraphics[width=0.45\textwidth]{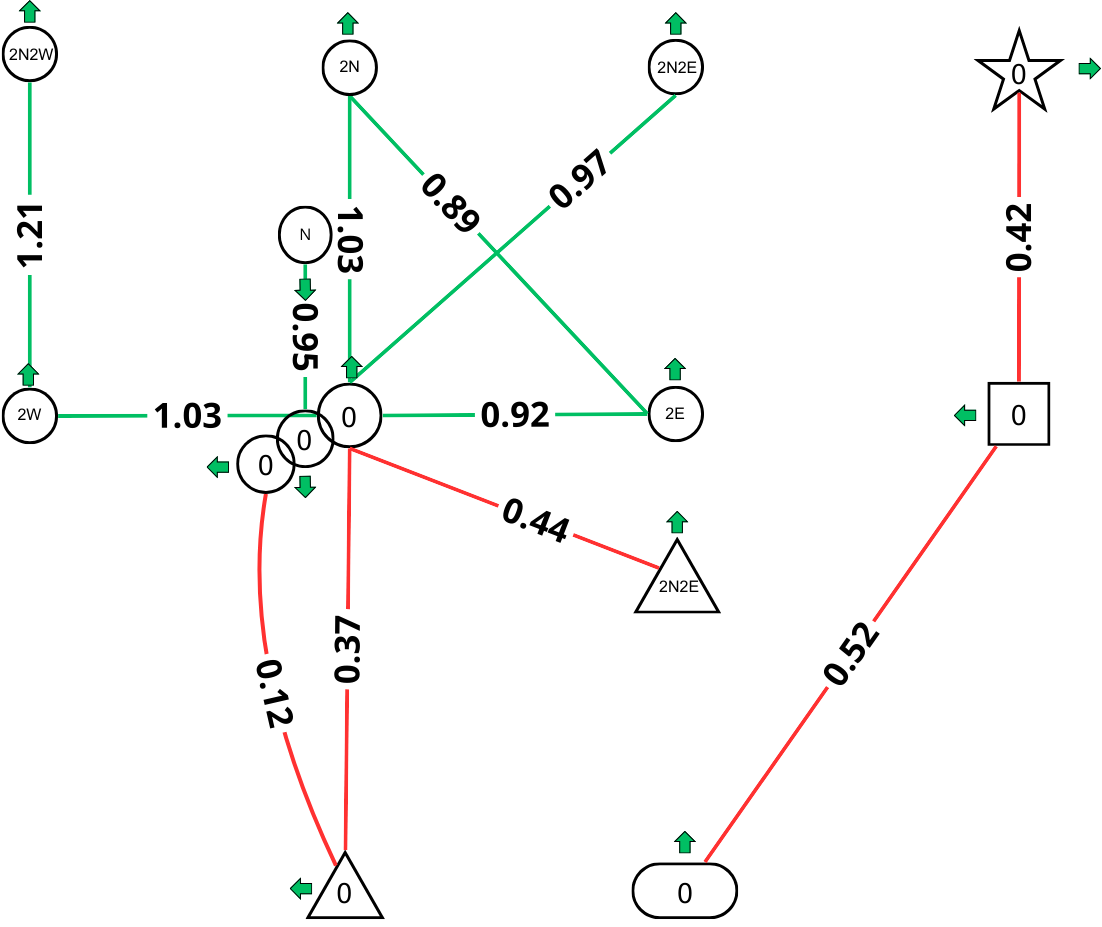}
    \caption{Distances across intra-class (green) and extra-class (red) holograms using the convolution-based score. The rounded element corresponds to the ``bullet'', the triangle to the ``PMN1'', the cylinder to the ``PMN4'', and the square to ``M-14''. N, E, W, and S corresponds to the shift of the corresponding object toward North, East, West, and South. The green arrows indicate the orientation of the objects.}
    \label{fig:pair-wise-conv}
\end{figure}

As shown in Figure \ref{fig:pair-wise-conv}, we have collected numerous scans of objects (Bullet, PMN-1, PMN-4, M-14) from different orientations (north, west, south, east) and with some controlled shifts, moving the objects one or two steps to the north, east, west, and south (N, E, W, S) and some combination of those (2N2W, 2N2E, etc.). In the figure, the rounded element corresponds to Bullet, the triangle to the PMN1, the cylinder to the PMN4, and the square to M-14. In particular, the orientations indicate the degree of rotation from one scan to another, and the shifts mean $2~cm$ to North, South, West, and East (N, S, W, E respectively). To evaluate the similarity, we have designed a convolution-based correlation metric between holographic scans where we measure the pairwise convolution score among the same and different object-soil holograms. As can be seen from Figure \ref{fig:pair-wise-conv}, the green line corresponds to the same object-soil holograms and their similarity score is between 0.90 and 1.2, with an average of 1.004. In contrast, the red ones correspond to different landmine-soil combinations and the scores vary from 0.12 to 0.5, with an average of 0.302.

With this setting, in order to find the optimal $\alpha$ coefficient, we aim at combining $H^{in}$ and $H^{out}$ with varying $\alpha$ and calculating the convolution-based score (correlation) with respect to the original natural scan. The original natural scans are acquired in the same weather condition of $H^{out}$ but with the landmine well-positioned under the soil. Once we have acquired all the natural scans, we remove all landmines from the soil and further scan the soil alone ($H^{out}$) and the landmine in-air ($H^{in}$) for the creation of the synthetic data.

\begin{figure}
    \hspace{-5mm}
    \includegraphics[width=0.5\textwidth]{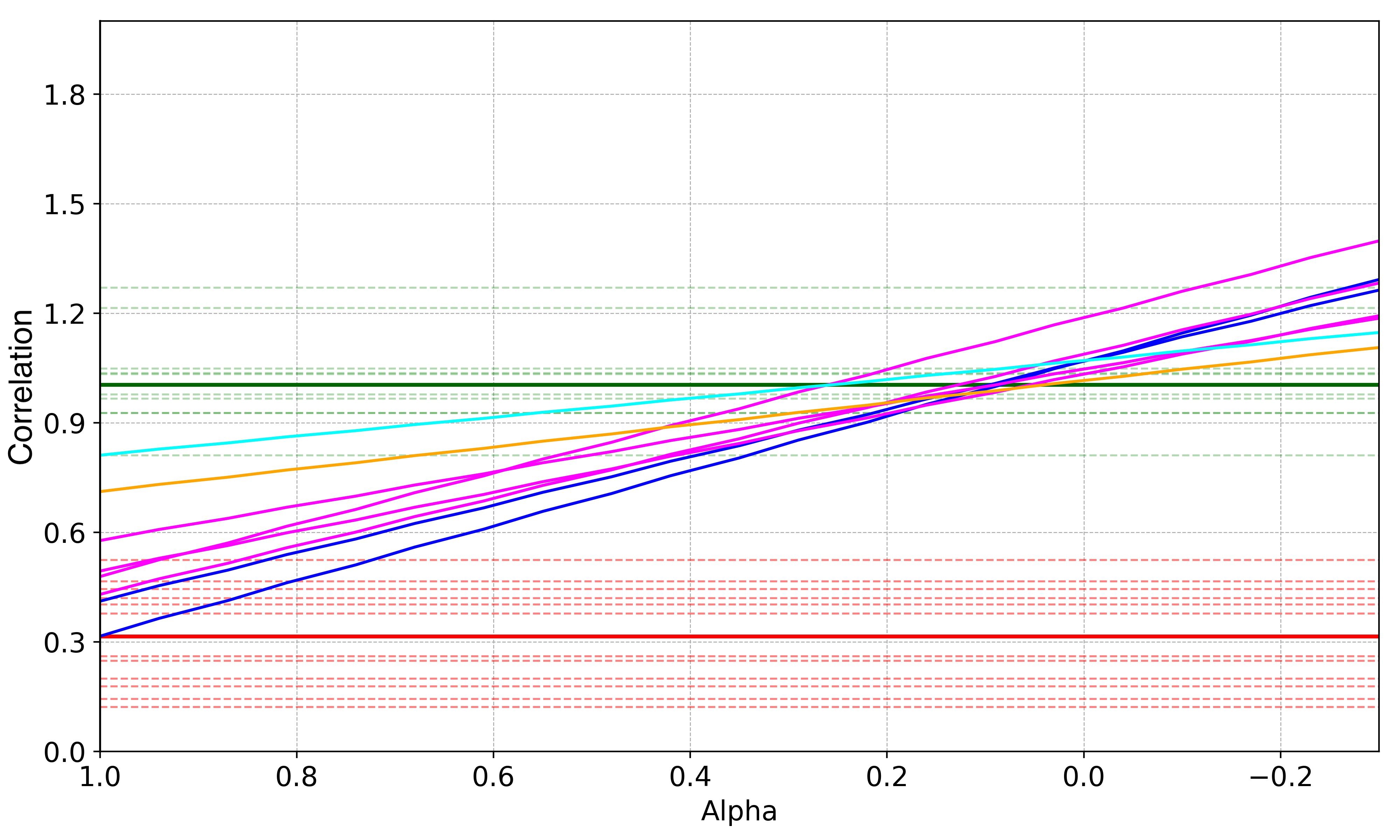}
    \caption{Convolution-based score (correlation) among the synthetic hologram with varying $\alpha$ and the natural scan in the same condition and setting. Every oblique line corresponds to one experimental setting, and the line colors correspond to Bullet (fuchsia), PMN-1 (cyan), PMN-4 (blue), and M-14 (yellow). The horizontal lines correspond to convolution-based scores for positive pairs (green dashes) or negative pairs (red dashes). The dark green and red indicate the means of these two sets of measurements.}
    \label{fig:correlation}
\end{figure}
Figure \ref{fig:correlation} reports the values of different synthetic holograms correlation to their respective natural scans, varying the $\alpha$ coefficient. The green-colored horizontal lines correspond to convolution-based correlations (obtained from Fig. \ref{fig:pair-wise-conv}), and the dark-green horizontal line corresponds to the optimal $\alpha$ coefficient obtained by averaging the per-sample coefficient, obtaining $1.004$ as the ``correlation alpha score''. With the same logic, the red horizontal lines correspond to the not-correlated convolution-based scores, and the darker red line to the uncorrelated mean. The other colors, as illustrated in the caption, correspond to the similarity between the mixed object-soil scans and the ground truth, at the variation of the alpha coefficient (x-axis). Optimal $\alpha$ corresponds to $\alpha^{*}=0.14$. In conclusion, by considering the dielectric permittivity of soil and air, and by empirically studying the correlation between synthetic and natural holographic images, we obtain similar results: i.e.~$\alpha^{*}=0.14$. Thus, the $\alpha$ coefficient used for hologram fusion is set to $0.14$ for all the objects. 

\subsection{Quantitative results}

\begin{table}[]
    \centering

    \resizebox{\linewidth}{!}{
    \begin{tabular}{l | c | ccc}
    Model        & Params & binary                & ternary               & multi                 \\ \midrule
    U-Net        & 35.8M  & 67,1 $\pm$ 2,6 & \textbf{52,67 $\pm$ 1,2} & 20,9 $\pm$ 2,4          \\
    ResNet       & 25.7M  & 55,7 $\pm$ 0,3          & 33,4 $\pm$ 0,15          & \textbf{24,7 $\pm$ 2,8} \\
    EfficientNet & 63.8 M & 40,3 $\pm$ 0,9          & 33,15 $\pm$ 0,4          & 8,1 $\pm$ 0,3           \\
    ViT          & 50.6 M & \textbf{69,3 $\pm$ 0,5}        & 44,9 $\pm$ 1,2          & 19,7 $\pm$ 1,3
    \end{tabular}
    
    }
    \caption{Experimental results for 2D classification tasks on the Buried Landmine Detection dataset. We report the f1-micro score for each model and task considering multiple runs, along with the number of parameters.}
    \label{tab:experiments-2D}
\end{table}

\begin{table}[]
    \centering
    
    \begin{tabular}{l|lll}
    Model         & binary                          & ternary                   & multi                  \\ \midrule
    U-Net 3D      & 60,6 $\pm$ 4,1                  & \textbf{47,67 $\pm$ 3,2}  & \textbf{19,9 $\pm$ 2,8}          \\
    ViT 3D        & \textbf{62,2 $\pm$ 3,2}         & 38,7 $\pm$ 2,5            & 16,5 $\pm$ 3,1
    \end{tabular}
    \caption{Experimental results for 3D classification tasks on the Buried Landmine Detection dataset. We report the f1-micro score for each model and task considering multiple runs, along with the number of parameters.}
    \label{tab:experiments-3D}
    \vspace{-2mm}
\end{table}

We conducted experiments to evaluate the performance of different deep learning model baselines on our proposed buried landmine detection dataset. A noteworthy trend in the tables is the trade-off between model complexity (measured by parameter count) and performance. Among the models evaluated, ResNet and EfficientNet are the smallest and largest, respectively. From Table \ref{tab:experiments-2D}, it can be observed that despite having less than half the parameters of EfficientNet, ResNet outperforms it in every classification task, especially in multi-class classification, which is the most challenging. A similar pattern is evident when comparing the best convolution-based model (U-Net) with the transformer-based model (ViT). Despite ViT being 50\% larger than U-Net, it only performs slightly better in the binary task and underperforms in all other tasks. This emphasizes the importance of balancing model complexity and performance in deep learning models designed for a specific task.
Moreover, the results suggest that 2D models may be more effective for detecting buried landmines than 3D models. This could be due to several factors, such as noise in the latter layer added during reconstruction phase or underfitting due to the complexity of the dataset and model dimensions. This finding contradicts the logical assumption that 3D models can capture more spatial information and have a better understanding of the shape and structure of buried objects compared to 2D models. Further investigation may be required to gain a better understanding of the capabilities of 3D models versus 2D models.

The effect of objects' sizes on detection/classification impacts the predictions, but that it seems that the material of the landmine (the ``PMN-4'' has a metallic ring, the ``Can'' is highly reflective, and so on) is more important.

\section{Conclusions}
In this paper we have presented a novel synthetic dataset supported by exhaustive empirical analysis, for buried landmines detection and classification, to foster the development of automated demining systems capable of dealing with buried landmines. The dataset has been created using a holographic subsurface radar and comprises 2D holographic images and 3D holographic inverted scans. To the best of our knowledge, this is the largest dataset of this type, in terms of a number of images and variety of mines and types of clutter objects.
Our experiments demonstrate that, while deep learning models show promise in buried landmine detection, there is still significant room for improvement to achieve the high levels of accuracy necessary for real-world use. In fact, even if some models achieve $F_{1-micro}$ scores as high as 69\%, these results in the context of landmine detection are unsatisfactory. By releasing this dataset we aim at fostering further research in the field, encouraging the development of more sophisticated deep learning architectures and training strategies that can enhance the detection performance. 

In our future work we plan to extend the dataset by adding other types of sensors, such as metal detectors and time-of-flight (ToF) cameras, to account for more classes of mines (e.g.~tripwire mines) and to better distinguish different types of mines. Additionally, we aim to improve computer vision approaches for the detection and classification of mines. Indeed, this paper only scratched the surface of the dataset's potential, as it focused solely on the classification task and employed only the real data (extracted amplitude components) from the complex data. As shown by the results obtained by the baselines, the problem is far from being solved and requires more effort from the computer vision community; we hope that the availability of this new large dataset will help the community to advance the research in this domain.

\section*{Acknowledgment}
This research work is funded by ASMARA project of Tuscany Region, and NATO SPS G-7563 project.

\ifCLASSOPTIONcaptionsoff
  \newpage
\fi

{\small
\bibliographystyle{IEEEtran}
\bibliography{main}
}

\begin{IEEEbiography}[{\includegraphics[width=1in,height=1.25in,clip,keepaspectratio]{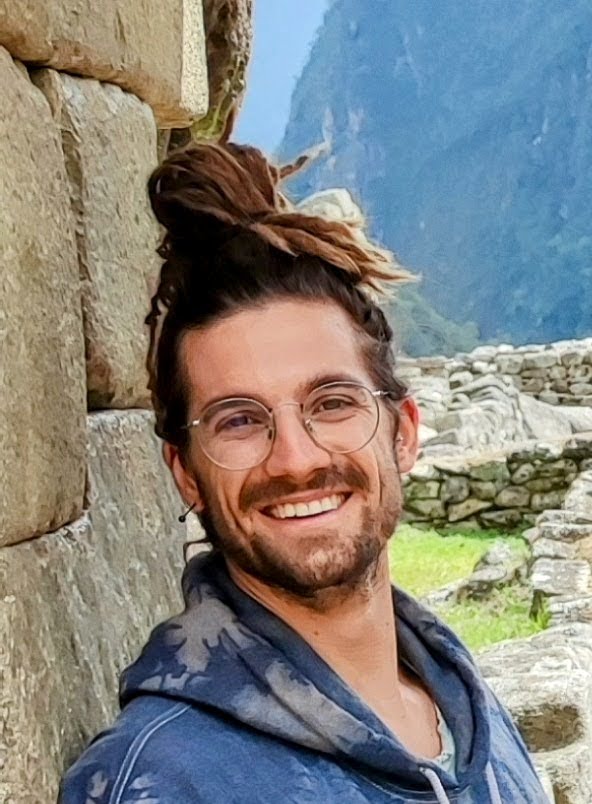}}]{Emanuele Vivoli}
Emanuele Vivoli is a third-year PhD student at the MICC Lab, University of Florence, and a second-year PhD student at the Computer Vision Center (CVC), Autonomous University of Barcelona. His research interests encompass multi-modal learning (vision and language integration), sensors, and robotic platforms, with a focus on multi-spectrum and multi-modal sensory data fusion. Throughout his academic career, Emanuele has made significant contributions as a first author to various conferences and journal articles. His work covers a diverse range of topics, including document and table analysis, understanding comics and manga, creating datasets and frameworks, and exploring multi-sensory data integration.
\end{IEEEbiography}

\begin{IEEEbiography}[{\includegraphics[width=1in,height=1.25in,clip,keepaspectratio]{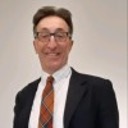}}]{Lorenzo Capineri}
Lorenzo Capineri. Scientific Qualification as Full Professor in Electronics. His current research activities are in the design of ultrasonic guided waves devices, buried objects detection with ground penetrating radar and holographic radar. He has worked on several research projects in collaboration with national (Gilardoni spa) and international industries (National Semiconductors, Texas Instruments, Marvell, Thales Alenia Space Italia), and research institutions: the Italian Research Council (CNR), the Italian Space Agency (ASI) and the European Space Agency (ESA), AEA Technology and UKAEA (England), ISTC (International Science and Technology Centre) (Moscow, Russia), European Commission Joint Research Centre (Ispra)  and NATO (Brussels, Belgium). He is coauthor of 7 Italian patents and coauthor of 4 book chapters and about 300 peer-revied scientific and technical papers. He is IEEE senior member since 2007 and member since 1983 and vice-president of the IEEE Italy Sensors Chapter. Co-chair of IWAGPR2015 conference and member of scientific and technical committee of IUS-IEEE, GPR, PIERS, URSI-GASS and IWAGPR conferences. Fellow of Electromagnetic Academy and Fellow British Institute of Non-Destructive Testing. 
\end{IEEEbiography}

\begin{IEEEbiography}[{\includegraphics[width=1in,height=1.25in,clip,keepaspectratio]{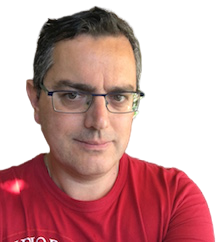}}]{Marco Bertini}
Marco Bertini is an Associate Professor in Computer Science at the University of Florence, Italy. He is the director of the Media Integration and Communication Center of the University of Florence. His interests are focused on multimedia and computer vision. On these subjects he has addressed semantic analysis, automatic content indexing, semantic retrieval and video quality improvement, applying these techniques to different domains among which cultural heritage. He is author of more than 30 journal papers and more than 150 peer-reviewed conference papers. He has been involved in 10 EU research projects as WP coordinator and researcher, among which IM3I, euTV, ORUSSI, UMETECH, AI4Media and ReInHerit. He has been general and program co-chair of several conferences on multimedia. He is co-founder of Small Pixels, an academic spin-off working on GenAI solutions to improve video quality and video compression.
\end{IEEEbiography}

\vfill

\end{document}